\documentclass[format=sigconf]{acmart}

\usepackage{amsmath,amssymb,amsfonts}
\usepackage{algorithmic}
\usepackage{graphicx}
\usepackage{textcomp}
\usepackage{xcolor}
\usepackage{xspace}
\usepackage{multicol}
\usepackage{multirow}
\usepackage{url}
\usepackage{listings}
\usepackage[utf8]{inputenc}
\usepackage{bm}
\usepackage{subcaption}
\usepackage{gensymb}
\usepackage{makecell}
\usepackage{etoolbox}
\usepackage{hyperref}
\usepackage{soul}
\usepackage{tikz}
\usepackage{balance}
\usepackage{color,soul}
\usepackage{flushend}

\DeclareRobustCommand\dottedred
{\tikz[baseline=0.0ex]\draw[red, densely dotted, line width=0.8pt] (0,0.1) -- (0.4,0.1);}
\DeclareRobustCommand\dottedblue {\tikz[baseline=0.0ex]\draw[blue, densely dotted, line width=0.8pt] (0,0.1) -- (0.4,0.1);}
\DeclareRobustCommand\dashedblack
{\tikz[baseline=0.0ex]\draw[black, densely dashed, line width=0.6pt] (0,0.1) -- (0.5,0.1);}
\DeclareRobustCommand\dashedred
{\tikz[baseline=0.0ex]\draw[red, dashdotted, line width=0.6pt] (0,0.1) -- (0.5,0.1);}

\newcommand{\legendsystem}{DarkneTZ}

\copyrightyear{2020} 
\acmYear{2020} 
\setcopyright{acmlicensed}
\acmConference[MobiSys '20]{The 18th Annual International Conference on Mobile Systems, Applications, and Services}{June 15--19, 2020}{Toronto, ON, Canada}
\acmBooktitle{The 18th Annual International Conference on Mobile Systems, Applications, and Services (MobiSys '20), June 15--19, 2020, Toronto, ON, Canada}
\acmPrice{15.00}
\acmDOI{10.1145/3386901.3388946}
\acmISBN{978-1-4503-7954-0/20/06}

\author[F. Mo]{Fan Mo}
\affiliation{Imperial College London}
\author[A. S. Shamsabadi]{Ali Shahin Shamsabadi}
\affiliation{Queen Mary University of London}
\author[K. Katevas]{Kleomenis Katevas}
\affiliation{Telef\'onica Research}
\author[S. Demetriou]{Soteris Demetriou}
\affiliation{Imperial College London}
\author[I. Leontiadis]{Ilias Leontiadis}
\affiliation{Samsung AI}
\author[A. Cavallaro]{Andrea Cavallaro}
\affiliation{Queen Mary University of London}
\author[H. Haddadi]{Hamed Haddadi}
\affiliation{Imperial College London}


\begin{document}

\title{DarkneTZ: Towards Model Privacy at the Edge using Trusted Execution Environments}

\begin{abstract}
\label{sec:abstract}

We present \emph{DarkneTZ}, a framework that uses an edge device's Trusted Execution Environment (TEE) in conjunction with model partitioning to limit the attack surface against Deep Neural Networks (DNNs). Increasingly, edge devices (smartphones and consumer IoT devices) are equipped with pre-trained DNNs for a variety of applications. This trend comes with privacy risks as models can leak information about their training data through effective membership inference attacks (MIAs).

We evaluate the performance of \emph{DarkneTZ}, including CPU execution time, memory usage, and accurate power consumption, using two small and six large image classification models. Due to the limited memory of the edge device's TEE, we partition model layers into more sensitive layers (to be executed inside the device TEE), and a set of layers to be executed in the untrusted part of the operating system. Our results show that even if a single layer is hidden, we can provide reliable model privacy and defend against state of the art MIAs, with only 3\% performance overhead. When fully utilizing the TEE, \emph{DarkneTZ} provides model protections with up to 10\% overhead.

\end{abstract}
\maketitle

\section{Introduction}
\label{sec:introduction}

Advances in memory and processing resources and the urge to reduce data transmission latency have led to a rapid rise in the deployment of various Deep Neural Networks (DNNs) on constrained edge devices (e.g.,~wearable, smartphones, and consumer Internet of Things (IoT) devices). Compared with centralized infrastructures (i.e.,~Cloud-based systems), hybrid and edge-based learning techniques enable methods for preserving users' privacy, as raw data can stay local~\cite{osia2018private}. Nonetheless, recent work demonstrated that local models still leak private information~\cite{yu2019differentially, hitaj2017deep, yeom2018privacy, melis2019exploiting, zhang2016understanding, zeiler2014visualizing, li2013membership, salem2018ml}. This can be used by adversaries aiming to compromise the confidentiality of the model itself or that of the participants in training the model~\cite{shokri2017membership, yeom2018privacy}. The latter, is part of a more general class of attacks, known as \emph{Membership Inference Attacks} (refer to as MIAs henceforth).
 
MIAs can have severe privacy consequences~\cite{li2013membership, salem2018ml} motivating a number of research works to focus on tackling them~\cite{abadi2016deep, mironov2017renyi, jayaraman2019relaxations}. Predominantly, such mitigation approaches rely on differential privacy~\cite{dwork2014algorithmic, rappor2014ulfar}, whose improvement in privacy preservation comes with an adverse effect on the model's prediction accuracy. 

We observe, that edge devices are now increasingly equipped with a set of software and hardware security mechanisms powered by processor (CPU) designs offering strong isolation guarantees. System designs such as Arm TrustZone can enforce memory isolation between an untrusted part of the system operating in a \textit{Rich Execution Environment} (REE), and a smaller trusted component operating in hardware-isolated \textit{Trusted Execution Environment} (TEE), responsible for security critical operations. If we could efficiently execute sensitive DNNs inside the trusted execution environments of mobile devices, this would allow us to limit the attack surface of models without impairing their classification performance. Previous work has demonstrated promising results in this space; recent advancements allow for high-performance execution of sensitive operations within a TEE~\cite{hunt2018chiron, hanzlik2018mlcapsule, tople2018privado, gu2018securing, tramer2018slalom}. These works have almost exclusively experimented with integrating DNNs in cloud-like devices equipped with Intel Software Guard eXtensions (SGX). However, this paradigm does not translate well to edge computing due to significant differences in the following three factors: attack surface, protection goals, and computational performance. The \emph{attack surface} on servers is exploited to steal a user's private data, while the adversary on a user's edge device focuses on compromising a model's privacy. Consequently, the \emph{protection goal} in most works combining deep learning with TEEs on the server (e.g.,~\cite{gu2018securing} and~\cite{hunt2018chiron}) is to preserve the privacy of a user's data during inference, while the protection on edge devices preserves both the model privacy and the privacy of the data used in training this model. Lastly, edge devices (such as IoT sensors and actuators) have \emph{limited computational resources} compared to cloud computing devices; hence we cannot merely use performance results derived on an SGX-enabled system on the server to extrapolate measurements for TEE-enabled embedded systems. In particular, blindly integrating a DNN in an edge device's TEE might not be computationally practical or even possible. We need a systematic measurement of the effects of such designs on edge-like environments.

Since DNNs follow a layered architecture, this can be exploited to partition a DNN, having a sequence of layers executed in the untrusted part of the system while hiding the execution of sensitive layers in the trusted, secure environment. We utilize the TEE (i.e., Arm TrustZone) and perform a unique layer-wise analysis to illustrate the privacy repercussions of an adversary on relevant neural network models on edge devices with the corresponding performance effects. To the best of our knowledge, we are the first to embark on examining to what extent this is feasible on resource-constrained mobile devices. Specifically, we lay out the following research question:

\textbf{RQ1:} \textit{Is it practical to store and execute a sequence of sensitive DNN's layers inside the TEE of an edge device?} 

To answer this question we design a framework, namely \emph{\legendsystem}, which enables an exhaustive layer by layer resource consumption analysis during the execution of a DNN model. \legendsystem{} partitions a model into a set of non-sensitive layers ran within the system's REE and a set of sensitive layers executed within the trusted TEE. We use \legendsystem{} to measure, for a given DNN---we evaluate two small and six large image classification models---the underlying system's CPU execution time, memory usage, and accurate power consumption for different layer partition choices. We demonstrate our prototype of \legendsystem{} using the Open Portable TEE (OP-TEE)\footnote{\url{https://www.op-tee.org/}} software stack running on a Hikey 960 board.\footnote{\url{https://www.96boards.org/product/hikey960/}} OP-TEE is compatible with the mobile-popular Arm TrustZone-enabled hardware, while our choice of hardware closely resembles common edge devices' capabilities~\cite{ying2018truz, park2019streambox}. Our results show that \legendsystem{} only has 10\% overhead when fully utilizing all available secure memory of the TEE for protecting a model's layers.

These results illustrate that REE-TEE partitions of certain DNNs can be efficiently executed on resource constrained devices. Given this, we next ask the following question:

\textbf{RQ2:} \textit{Are such partitions useful to both effectively and efficiently tackle realistic attacks against DNNs on mobile devices?} 

To answer this question, we develop a threat model considering state of the art MIAs against DNNs. We implement the respective attacks and use \legendsystem{} to measure their effectiveness (adversary's success rate) for different model partition choices. We show that by hiding a single layer (the output layer) in the TEE of a resource-constrained edge device, the adversary's success rate degrades to random guess while (a) the resource consumption overhead on the device is negligible (3\%) and (b) the accuracy of inference remains intact. We also demonstrate the overhead of fully utilizing TrustZone for protecting models, and show that \legendsystem{} can be an effective first step towards achieving hardware-based model privacy on edge devices.

\vspace{5pt}\noindent\textbf{Paper Organisation.} The rest of the paper is organized as follows: Section~\ref{sec:related_work} discusses background and related work and Section~\ref{sec:framework} presents the design and main components of \legendsystem{}. Section~\ref{sec:setup} provides implementation details and describes our evaluation setup (our implementation is available online\footnote{\url{https://github.com/mofanv/darknetz}}), while Section~\ref{sec:validation} presents our system performance and privacy evaluation results. Lastly, Section~\ref{sec:discussion} discusses further performance and privacy implications that can be drawn from our systematic evaluation and we conclude on Section~\ref{sec:conclusion}.
\section{Background and Related Work}
\label{sec:related_work}

%%%%%%%%%%%%%%  subsection  %%%%%%%%%%%%%%
\subsection{Privacy risks of Deep Neural Networks}
\label{sec:relatedwork_mia}

\noindent\textbf{Model privacy risks.} With successful training (i.e.,~the model converging to an optimal solution), a DNN model ``memorizes" features of the input training data~\cite{yeom2018privacy, radhakrishnan2018downsampling} (see \cite{8666641, lecun2015deep} for more details on deep learning), which it can then use to recognize unseen data exhibiting similar patterns. However, models have the tendency to include more specific information of the training dataset unrelated to the target patterns (i.e.,~the classes that the model aims to classify)~\cite{yeom2018privacy, caruana2001overfitting}.

Moreover, each layer of the model memorizes different information about the input. Yosinki et al.~\cite{yosinski2014transferable} found that the first layers (closer to the input) are more transferable to new datasets than the last layers. That is, the first layers learn more \textit{general} information (e.g.,~ambient colors in images), while the last layers learn information that is more \textit{specific} to the classification task (e.g.,~face identity). The memorization difference per layer has been verified both in convolutional layers~\cite{zeiler2014visualizing, yosinski2015understanding} and in generative models~\cite{zhao2017learning}. Evidently, an untrusted party with access to the model can leverage the memorized information to infer potentially sensitive properties about the input data which leads to severe privacy risks.

\vspace{5pt}\noindent\textbf{Membership inference attack (MIA)}. MIAs form a possible attack on devices which leverage memorized information on a models' layers to determine whether a given data record was part of the model's training dataset~\cite{shokri2017membership}. In a \emph{black-box MIA}, the attacker leverages models' outputs (e.g.,~confidence scores) and auxiliary information (e.g.,~public datasets or public prediction accuracy of the model) to train shadow models or classifiers without accessing internal information of the model~\cite{shokri2017membership, yeom2018privacy}. However, in a \emph{white-box} MIA, the attacker utilizes the internal knowledge (i.e.,~gradients and activation of layers) of the model in addition to the model's outputs to increase the effectiveness of the attack~\cite{nasr2018comprehensive}. It is shown that the last layer (model output) has the highest membership information about the training data~\cite{nasr2018comprehensive}. We consider a white-box adversary as our threat model, as DNNs are fully accessible after being transferred from the server to edge devices~\cite{xu2019first}. In addition to this, a white-box MIA is a \emph{stronger} adversary than a black-box MIA, as the information the adversary has access to in a black-box attack is a subset of that used in a white-box attack.

%%%%%%%%%%%%%%  subsection  %%%%%%%%%%%%%%
\subsection{Deep learning in the TEE}

\noindent\textbf{Trusted execution environment (TEE).} A TEE is a trusted component which runs in parallel with the untrusted Rich operating system Execution Environment (REE) and is designed to provide safeguards for ensuring the confidentiality and integrity of its data and programs. This is achieved by establishing an isolated region on the main processor, and both hardware and software approaches are utilized to isolate this region. The chip includes additional elements such as unchangeable private keys or secure bits during manufacturing, which helps ensure that untrusted parts of the platform (even privileged OS or hypervisor processes) cannot access TEE content~\cite{costan2016intel, arm2009security}.

In addition to strong security guarantees, TEEs also provide better computational performance than existing software protections, making it suitable for computationally-expensive deep learning tasks. For example, advanced techniques such as fully homomorphic encryption enable operators to process the encrypted data and models without decryption during deep learning, but significantly increase the computation cost~\cite{naehrig2011can, acar2018survey}. Conversely, TEE protection only requires additional operations to build the trusted environment and the communication between trusted and untrusted parts, so its performance is comparable to normal executions in an untrusted environment (e.g.,~an OS).

\vspace{5pt}\noindent\textbf{Deep learning with TEEs.} Previous work leveraged TEEs to protect deep learning models. Apart from the unique attack surface and thus protection goals we consider, these also differ with our approach in one more aspect: they depend on an underlying computer architecture which is more suitable for cloud environments. Recent work has suggested executing a complete deep learning model in a TEE~\cite{costan2016intel}, where during training, users' private data is transferred to the trusted environment using trusted paths. This prevents the host Cloud form eavesdropping on the data~\cite{ohrimenko2016oblivious}. Several other studies improved the efficiency of TEE-resident models using Graphics Processing Units (GPU)~\cite{tramer2018slalom}, multiple memory blocks~\cite{hunt2018chiron}, and high-performance ML frameworks~\cite{hynes2018efficient}. More similar to our approach, Gu et al.~\cite{gu2018securing} partitioned DNN models and only enclosed the first layers in an SGX-powered TEE to mitigate input information disclosures of real-time fed device user images. In contrast, membership inference attacks we consider, become more effective by accessing information in the last layers. All these works use an underlying architecture based on Intel's SGX, which is not suitable for edge devices. Edge devices usually have chips designed using Reduced Instruction Set Computing (RISC), peripheral interfaces, and much lower computational resources (around 16 mebibytes (MiB) memory for TEE)~\cite{ekberg2014untapped}. Arm's TrustZone is the most widely used TEE implementation in edge devices. It involves a more comprehensive trusted environment, including the security extensions for the AXI system bus, processors, interrupt controller, TrustZone address space controller, etc. Camera or voice input connected to the APB peripheral bus can be controlled as a part of the trusted environment by the AXI-to-APB bridge. Utilizing TrustZone for on-device deep learning requires more developments and investigations because of its different features compared to SGX.

%%%%%%%%%%%%%%  subsection  %%%%%%%%%%%%%%
\subsection{Privacy-preserving methods}

An effective method for reducing the memorization of private information of training data in a DNN model is to avoid \emph{overfitting} via imposing constraints on the parameters and utilizing dropouts~\cite{shokri2017membership}. \emph{Differential Privacy} (DP) can also obfuscate the parameters (e.g.,~adding Gaussian noise to the gradients) during training to control each input's impact on them~\cite{abadi2016deep, yu2019differentially}. However, DP may negatively affect the utility (i.e.,~the prediction accuracy) if the noise is not carefully designed~\cite{rahman2018membership}. In order to obfuscate private information only, one could apply methods such as generative neural networks~\cite{xu2019ganobfuscator} or adversarial examples~\cite{jia2019memguard} to craft noises for one particular data record (e.g., one image), but this requires additional computational resources which are already limited on edge devices.

\vspace{5pt}\noindent\textbf{Server-Client model partition.} General information processed in the first layers ~\cite{yosinski2014transferable} during forward propagation of deep learning often includes more important indicators for the inputs than those in the last layers (which is opposite to membership indicators), since reconstructing the updated gradients or activation of the first layers can directly reveal private information of the input~\cite{aono2018privacy, dosovitskiy2016inverting}. Based on this, hybrid training models have been proposed which run several first layers at the client-side for feature extraction and then upload these features to the server-side for classification~\cite{osia2017hybrid}. Such partition approaches delegate parts of the computation from the servers to the clients, and thus, in these scenarios, striking a balance between privacy and performance is of paramount importance.

Gu et al.~\cite{gu2018securing} follow a similar layer-wise method and leverage TEEs on the cloud to isolate the more private layers. Clients' private data are encrypted and then fed into the cloud TEE so that the data and first several layers are protected. This method expands the clients' trusted boundary to include the server's TEE and utilizes an REE-TEE model partition at the server which does not significantly increase clients' computation cost compared to running the first layers on client devices. To further increase training speed, it is also possible to transfer all linear layers outside a cloud's TEE into an untrusted GPU~\cite{tramer2018slalom}. All these partitioning approaches aim to prevent leakage of private information of users (to the server or others), yet \emph{do not} prevent leakage from trained models when models are executed on the users' edge devices.
\section{\legendsystem}
\label{sec:framework}

\begin{figure}[t!]
    \centering
    \includegraphics[width=\columnwidth]{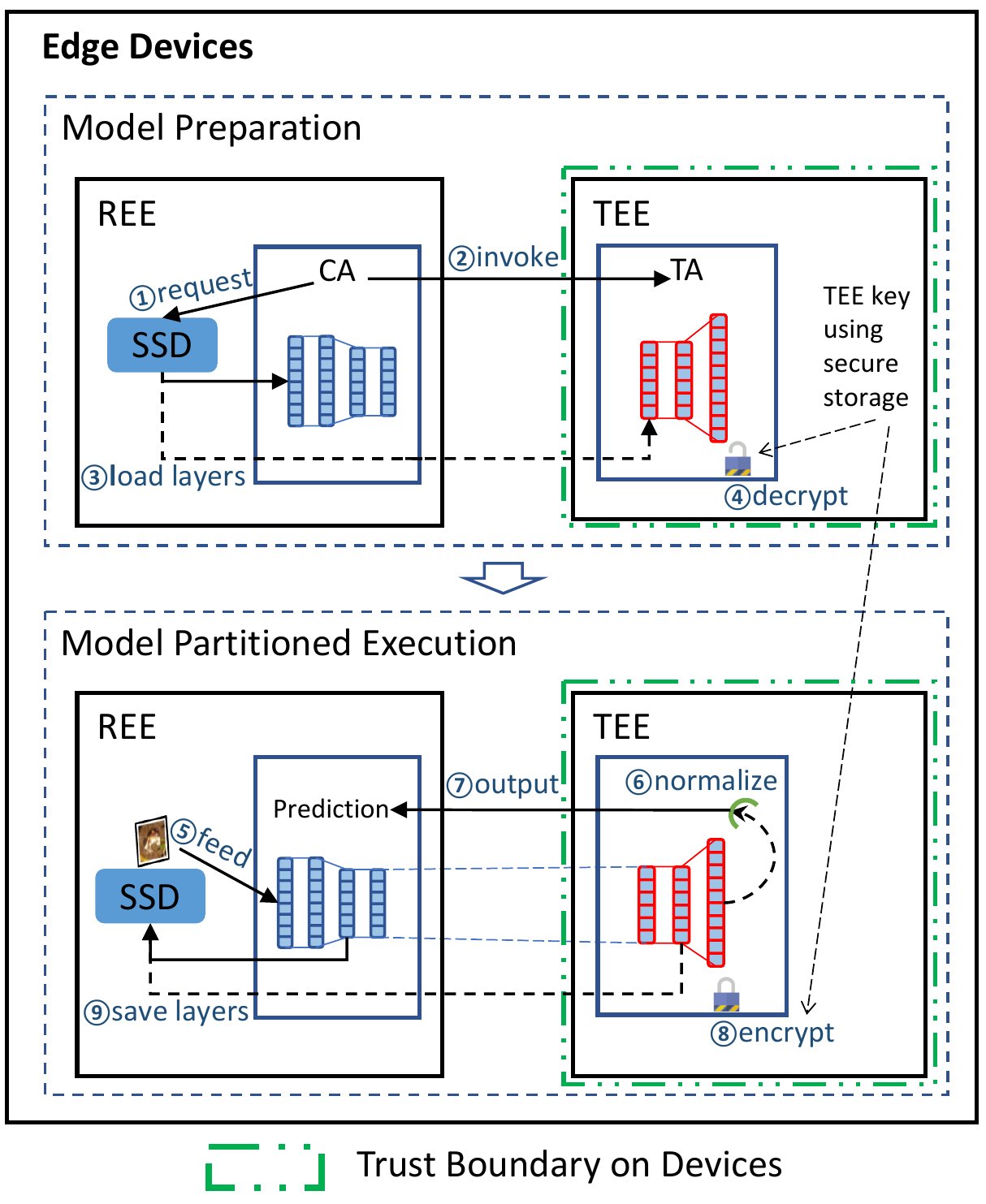}
    \caption{\legendsystem{} uses on-device TEE to protect a set of layers of a deep neural network for both inference and fine-tuning. (Note: The trusted compute base---or trust boundary---for the model owner on edge devices is the TEE of the device).}
    \label{fig:scenario}
\end{figure}

We now describe \legendsystem{}, a framework for preserving DNN models' privacy on edge devices. We start with the threat model which we focus on in this paper.

%%%%%%%%%%%%%%  subsection  %%%%%%%%%%%%%%
\subsection{Threat Model}
We consider an adversary with full access to the REE of an edge device (e.g.,~the OS) on edge devices: this could be the actual user, malicious third-party software installed on the devices, or a malicious or compromised OS. We only trust the TEE of an edge device to guarantee the integrity and confidentiality of the data and software in it. In particular, we assume that a DNN model is pre-trained using private data from the server or other participating nodes. We assume the model providers can fully guarantee the model privacy during training on their servers by utilizing existing protection methods~\cite{ohrimenko2016oblivious} or even by training the model offline, so the model can be secret provisioned to the user devices without other privacy issues.

%%%%%%%%%%%%%%  subsection  %%%%%%%%%%%%%%
\subsection{Design Overview}

\legendsystem{} design aims at mitigating attacks on on-device models by protecting layers and the output of the model with low cost by utilizing an on-device TEE. It should be compatible with edge devices. That is, it should integrate with TEEs which can run on hardware technologies that can be found on commodity edge devices (e.g. Arm TrustZone), use standard TEE system architectures and corresponding APIs.

We propose \legendsystem, illustrated in Figure~\ref{fig:scenario}, a framework that enables DNN layers to be partitioned as two parts to be deployed respectively into the REE and TEE of edge devices. \legendsystem{} allows users to do inference with or fine-tuning of a model seamlessly---the partition is transparent to the user---while at the same time considers the privacy concerns of the model's owner. Corresponding Client Application (CA) and Trusted Application (TA) perform the operations in REE and TEE, respectively. Without loss of generality, \legendsystem{}'s CA runs layers $1$ to $l$ in the REE, while its TA runs layers $l+1$ to $L$ located in the TEE during fine-tuning or inference of a DNN. This DNN partitioning can help the server to mitigate several attacks such as MIAs~\cite{nasr2018comprehensive, mo2019towards}, as the last layers have a higher probability of leaking private information about training data (see Section~\ref{sec:related_work}).

\legendsystem{} expects sets of layers to be pre-provisioned in the TEE by the analyst (if the framework is used for offline measurements) or by the device OEM if a version of \legendsystem{} is implemented on consumer devices. Note that in the latter case, secret provisioning of sensitive layers can also be performed over the air, which might be useful when the sensitive layer selection needs to be dynamically determined and provisioned to the edge device after supply. In this case, one could extend \legendsystem{} to follow a variation of the SIGMA secure key exchange protocol ~\cite{krawczyk2003sigma}, modified to include remote attestation, similar to~\cite{zhao2019sectee}. SIGMA is provably secure and efficient. It guarantees perfect forward secrecy for the session key (to defend against replay attacks) while its use of message authentication codes ensures server and client identity protection. Integrating remote attestation guarantees that the server provisions the model to a non-compromised edge device.

%%%%%%%%%%%%%%  subsection  %%%%%%%%%%%%%%
\subsection{Model Preparation}

Once the model is provisioned, the CA requests the layers from devices (e.g.,~solid-state disk drive (SSD)) and invokes the TA. The CA will first build the DNN architecture and load the parameters of the model into normal memory (i.e.,~non-secure memory) to process all calculations and manipulations of the non-sensitive layers in the REE. When encountering (secretly provisioned) encrypted layers need to be executed in the TEE, which is determined by the model owner's setting, the CA passes them to the TA. The TA decrypts these layers using a key that is securely stored in the TEE (using secure storage), and then it runs the more sensitive layers in the TEE's secure memory. The secure memory is indicated by one additional address bit introduced to all memory system transactions (e.g.,~cache tags, memory, and peripherals) to block non-secure access~\cite{arm2009security}. At this point, the model is ready for fine-tuning and inference.

%%%%%%%%%%%%%%  subsection  %%%%%%%%%%%%%%
\subsection{DNN Partitioned Execution}
\label{sec:dnn_partitioned_exe}

The \emph{forward pass} of both inference and fine-tuning passes the input $\mathbf{a}^0$ to the DNN to produce activation of layers until the last layer, i.e., layer $l$'s activation is calculated by $\mathbf{a}^l = f(\mathbf{w}^l \mathbf{a}^{l-1}+\mathbf{b}^l)$, where $\mathbf{w}^l$ and $\mathbf{b}^l$ are weights and biases of this layer, $\mathbf{a}^{l-1}$ is activation of its previous layer and $f$ is the non-linear activation function. Therefore, after the CA processes its inside layers from $1$ to $l$, it invokes a command to transfer the outputs (i.e.,~activation) of layer $l$ (i.e.,~the last layer in the CA) to the secure memory through a \emph{buffer} (in shared memory). The TA switches to the \emph{forward\_net\_TA} function corresponding to the invoked command to receive parameters (i.e., outputs/activation) of layer $l$ and processes the following forward pass of the network (from layer $l+1$ to layer $L$) in the TEE. In the end, outputs of the last layer are first normalized as $\hat{\mathbf{a}}^L$ to control the membership information leakage and are returned via shared memory as the prediction results.

The \emph{backward pass} of fine-tuning computes gradients of the loss function $\mathcal{L}(\mathbf{a}^L,y)$ with respect to each weight $\mathbf{w}^{l}$ and bias $\mathbf{b}^l$, and updates the parameters of all layers, $\{\mathbf{w}^{l}\}^L_{l=1}$ and $\{\mathbf{b}^{l}\}^L_{l=1}$ as 
$\mathbf{w}^l = \mathbf{w}^l - \eta \frac{\partial \mathcal{L}(\mathbf{a}^L,y)}{\partial \mathbf{w}^{l}}$ and $\mathbf{b}^l = \mathbf{b}^l - \eta \frac{\partial \mathcal{L}(\mathbf{a}^L,y)}{\partial \mathbf{b}^{l}}$, where $\eta$ is a constant called the learning rate and $y$ is the desired output (i.e.,~called label).
The TA can compute the gradient of the loss function by receiving $y$ from CA and back propagate it to the CA in order to update all the parameters. In the end, to save the fine-tuned model on devices, all layers in the TA are encrypted and transferred back to the CA.
\section{Experiment Settings}
\label{sec:setup}

%%%%%%%%%%%%%%  subsection  %%%%%%%%%%%%%%
\subsection{Models and Datasets}
\label{sec:model_dataset}

We first use two popular DNNs, namely AlexNet and VGG-7, to measure the system's performance. AlexNet has five convolutional layers (i.e.,~with kernel size 11, 5, 3, 3, and 3) followed by a fully-connected and a softmax layer, and VGG-7 has eight layers (i.e.,~seven convolutional layers with kernel size 3, followed by a fully-connected layer). Both AlexNet and VGG-7 use ReLU (Rectifier Linear Unit) activation functions for all convolutional layers. The number of neurons for AlexNet's layers is 64, 192, 384, 256, and 256, while the number of neurons for VGG-7's layers is 64, 64, 124, 124, 124, 124, and 124. We train the networks and conduct inference on CIFAR-100 and ImageNet Tiny. We use image classification datasets, as a recent empirical study shows that the majority of smartphone applications (70.6\%) that use deep learning are for image processing~\cite{xu2019first}. Moreover, the state of the art MIA we are considering is demonstrated against such datasets~\cite{nasr2018comprehensive}. CIFAR-100 includes 50k training and 10k test images of size $32 \times 32 \times 3$ belonging to 100 classes. ImageNet Tiny is a simplified ImageNet challenge that has 100k training and 10k test images of size $64 \times 64 \times 3$ belonging to 200 classes. 

In addition to this, we use six available DNNs (Tiny Darknet (4 megabytes (MB)), Darknet Reference (28MB), Extraction~\cite{szegedy2015going} (90MB), Resnet-50~\cite{he2016deep} (87MB), Densenet-201~\cite{huang2017densely} (66MB), and Darknet-53-448 (159MB)) pre-trained on the original ImageNet~\cite{deng2009imagenet} dataset to measure \legendsystem's performance during \emph{inference}. All pre-trained models can be found online\footnote{\url{https://pjreddie.com/darknet/imagenet/}}. ImageNet has 1000 classes, and consequently, these DNN models' last layers occupy larger memory that can exceed the TEE's limits, compared to models with 100/200 classes. Therefore, for these six models, we only evaluate the condition that their \emph{last layer} is in the TEE.

To evaluate the defence's effectiveness against MIAs, we use the same models as those used in the demonstration of the attack\cite{nasr2018comprehensive} (AlexNet, VGG-7, and ResNet-110). This ResNet with 110 depth is an existing network architecture that has three blocks (each has 36 convolutional layers) in the middle and another one convolutional layer at the beginning and one fully connected layer at the end~\cite{he2016deep}. We use published models trained (with 164 epochs) on CIFAR-100~\cite{cifar100} online\footnote{\url{https://github.com/bearpaw/pytorch-classification}}. We also train three models on ImageNet Tiny\footnote{\url{https://tiny-imagenet.herokuapp.com/}} with 300 epochs as target models (i.e.,~victim models during attacks). Models with the highest valid accuracy are used after training. We follow \cite{nasr2018comprehensive}'s methodology, and all training and test datasets are split to two parts with equal sizes randomly so that the MIA model learns both \emph{Member} and \emph{Non-member} images. For example, 25K of training images and 5K of test CIFAR-100 images are chosen to train the MIA model, and then the model's test precision and recall are evaluated using 5K of training images and 5K of test images in the rest of CIFAR-100 images.

%%%%%%%%%%%%%%  subsection  %%%%%%%%%%%%%%
\subsection{Implementation and Evaluation Setup}

We develop an implementation based on the  Darknet~\cite{darknet13} DNN library. We chose this particular library because of its high computational performance and small library dependencies which fits within the limited secure memory of the TEE. We run the implementation on Open Portable TEE (OP-TEE), which provides the software (i.e., operating systems) for an REE and a TEE designed to run on top of Arm TrustZone-enabled hardware.

For TEE measurements, we focus on the performance of deep learning since secret provisioning only happens once for updating the model from severs. We implement 128-bit AES-GCM for on-device secure storage of sensitive layers. We test our implementation on a Hikey 960 board, a widely-used device~\cite{ying2018truz,akowuah2018protecting,dong2018tzdks,brasser2019sanctuary} that is promising to be comparable with mobile phones (and other existing products) due to its Android open source project support. The board has four ARM Cortex-A73 cores and four ARM Cortex-A53 cores (pre-configured to 2362MHz and 533MHz, respectively, by the device OEM), 4GB LPDDR4 SDRAM, and provides 16MiB secure memory for trusted execution, which includes 14MiB for the TA and 2MiB for TEE run-time. Another 2MiB shared memory is allocated from non-secure memory. As the Hikey board adjusts the CPU frequency automatically according to the CPU temperature, we decrease and fix the frequency of Cortex A73 to 903MHz and keep the frequency of Cortex A53 as 533Mhz. During experiments we introduce a 120 seconds system sleep per trial to make sure that the CPU temperature begins under $40\degree C$ to avoid underclocking.

Edge devices suffer from limited computational resources, and as such, it is paramount to measure the efficiency of deep learning models when partitioned to be executed partly by the OS and partly by the TEE. In particular we monitor and report CPU execution time (in seconds), memory usage (in megabytes), and power consumption (in watts) when the complete model runs in the REE (i.e.,~OS) and compare it with different partitioning configurations where more sensitive layers are kept within the TEE. CPU execution time is the amount of time that the CPU was used for deep learning operations (i.e.,~fine-tuning or inference). Memory usage is the amount of the mapping that is currently resident in the main memory (RAM) occupied by our process for deep learning related operations. Power consumption is the electrical energy consumption per unit time that was required by the Hikey board.

More specifically, we utilized the REE's \texttt{/proc/self/status} for accessing the process information to measure the CPU execution time and memory usage of our implementation. CPU execution time is the amount of time for which the CPU was used for processing instructions of software (as opposed to wall-clock time which includes input/output operations) and is further split into (a) time in user mode and (b) time in kernel mode. The REE kernel time captures together (1) the time spent by the REE’s kernel and (2) the time spent by the TEE (including both while in user mode and kernel mode). This kernel time gives us a direct perception of the overhead when including TEEs for deep learning versus using the same REE without a TEE's involvement.

Memory usage is represented using \emph{resident set size (RSS) memory} in the \emph{REE}, but the memory occupied in the TEE is not counted by the RSS since the REE does not have access to gather memory usage information of the TEE. The TEE is designed to conceal this sensitive information (e.g., both CPU time and memory usage); otherwise, the confidentiality of TEE contents would be easily breached by utilizing side-channel attacks~\cite{wang2017leaky}. To overcome this, we trigger an abort from the TEE after the process runs stably (memory usage tends to be fixed) to obtain the memory usage of the TEE.

To accurately measure the power consumption, we used Monsoon High Voltage Power Monitor,\footnote{\url{https://www.msoon.com/}} a high-precision power metering hardware capable of measuring the current consumed by a test device with a voltage range of 0.8V to 13.5V and up to 6A continuous current. We configured it to power the Hikey board using the required 12V voltage while recording the consumed current in a $50Hz$ sampling rate.

For conducting the MIA, we use a machine with 4 Intel(R) Xeon(R) E5-2620 CPUs (2.00GHz), an NVIDIA QUADRO RTX 6000 (24GB), and 24GB DDR4 RAM. Pytorch v1.0.1~\cite{paszke2017automatic} is used as the DNN library.

%%%%%%%%%%%%%%  subsection  %%%%%%%%%%%%%%
\subsection{Measuring Privacy in MIAs}
\label{sec:measure_mia}

We define the adversarial strategy in our setting based on state-of-the-art white-box MIAs which observe the behavior of all components of the DNN model~\cite{nasr2018comprehensive}. White-box MIAs can achieve higher accuracy of distinguishing whether one input sample is presented in the private training dataset compared to black-box MIAs since the latter only have access to the models' output~\cite{yeom2018privacy, shokri2017membership}. Besides, white-box MIAs are also highly possible in on-device deep learning, where a model user can not only observe the output, but also observe fine-grained information such as the values of the cost function, gradients, and activation of layers.

We evaluate the membership information exposure of a set of the target model's layers by employing the white-box MIA~\cite{nasr2018comprehensive} on these layers. The attacker feeds the target data to the model and leverages all possible information in the white-box setting including activation of all layers, model's output, loss function, and the gradients of the loss function with respect to the parameter of each layer. It then separately analyses each information source by extracting features from the activation of each layer, the model's output and the loss function via fully connected neural networks with one hidden layer, while using convolutional neural networks for the gradients. All extracted features are combined in a global feature vector that is later used as an input for an inference attack model. The attack model predicts a single value (i.e.,~Member or Non-member) that represents the membership information of the target data (we refer the interested readers to \cite{nasr2018comprehensive} for a detailed description of this MIA). We use the test accuracy of the MIA model trained on a set of layers to represent the advantage of adversaries as well as the sensitivity of these layers.

To measure the privacy risk when part of the model is in TEE, we conduct this MIA on our target model in two different settings: (i) starting from the first layer, we add the later layers one by one until the end of the network, and (ii) starting from the last layer we add the previous layers one by one until the beginning of the network. However, the available information of one specific layer during the \textit{fine-tuning phase} and that during the \textit{inference phase} are different when starting from the first layers. Inference only has a forward propagation phase which computes the activation of each layer. During fine-tuning and because of the backward propagation, in addition to the activation, gradients of layers are also visible. In contrast to that, attacks starting from the last layers can observe the same information in both inference and fine-tuning since layers' gradients can be calculated based on the cost function. Therefore, in setting (i), we utilize activation, gradients, and outputs. In setting (ii), we only use the activation of each layer to evaluate inference and use both activation and gradients to evaluate fine-tuning, since the outputs of the model (e.g., confidence scores) are not accessible in this setup.

\section{Evaluation Results}
\label{sec:validation}

In this Section we first evaluate the efficiency of \legendsystem{} when protecting a set of layers in the TrustZone to answer \textbf{RQ1}. To evaluate system efficiency, we measure \textbf{CPU execution time}, \textbf{memory usage}, and \textbf{power consumption} of our implementation for both training and inference on AlexNet and VGG-7 trained on two datasets. We protect the last layers (starting from the output) since they are more vulnerable to attacks (e.g.,~MIAs) on models. The cost layer (i.e.,~the cost function) and the softmax layer are considered as a separate layer since they contain highly sensitive information (i.e.,~confidence scores and cost function). Starting from the last layer, we include the maximum number of layers that the TrustZone can hold. To answer \textbf{RQ2}, we use the \textbf{MIA success rate}, indicating the membership probability of target data (the more \legendsystem{} limits this, the stronger the privacy guarantees are). We demonstrate the effect on performance and discuss the trade-off between performance and privacy using MIAs as one example.

%%%%%%%%%%%%%%  subsection  %%%%%%%%%%%%%%
\subsection{CPU Execution Time}

%%%
\begin{figure*}[t!]
\centering
     \begin{subfigure}{1.13\columnwidth}
     \includegraphics[scale=0.65]{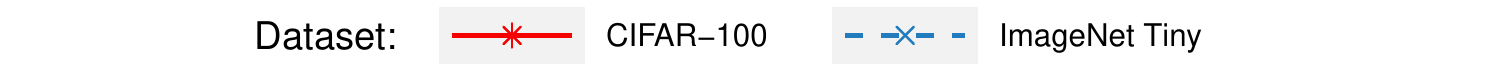}
     \end{subfigure} %
     \\
    \begin{subfigure}{1\columnwidth}
     \includegraphics[scale=0.65]{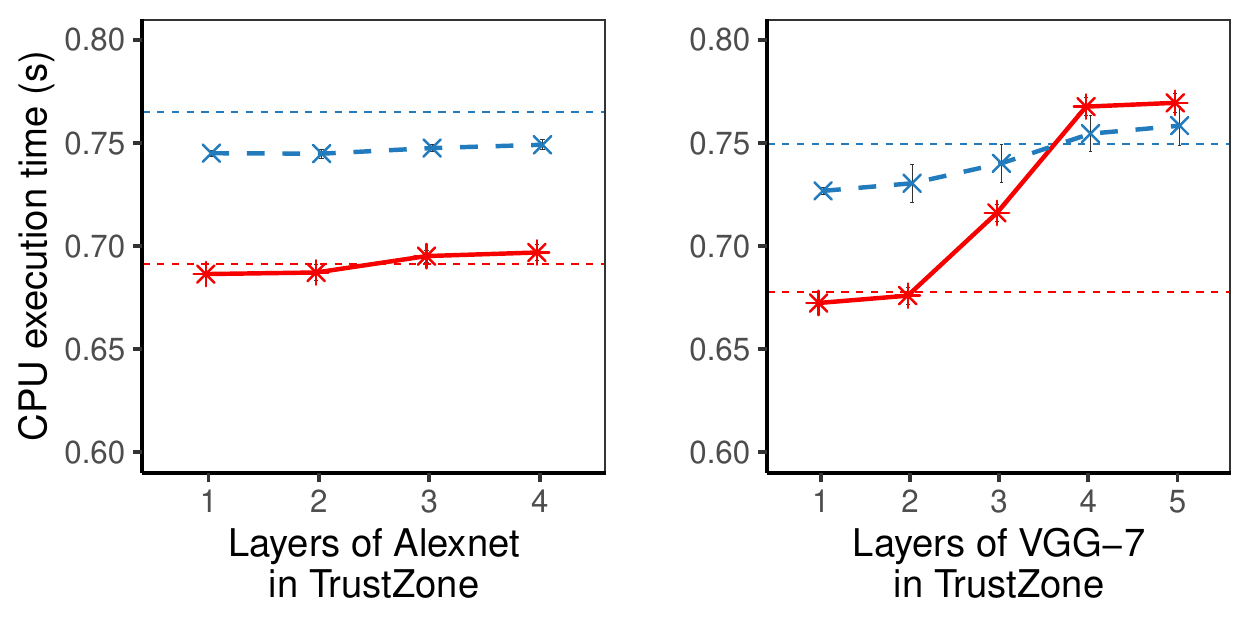}
     \caption{CPU time of training}\label{fig:SM_exe_time_train}
   \end{subfigure}%
    \begin{subfigure}{1\columnwidth}
     \includegraphics[scale=0.65]{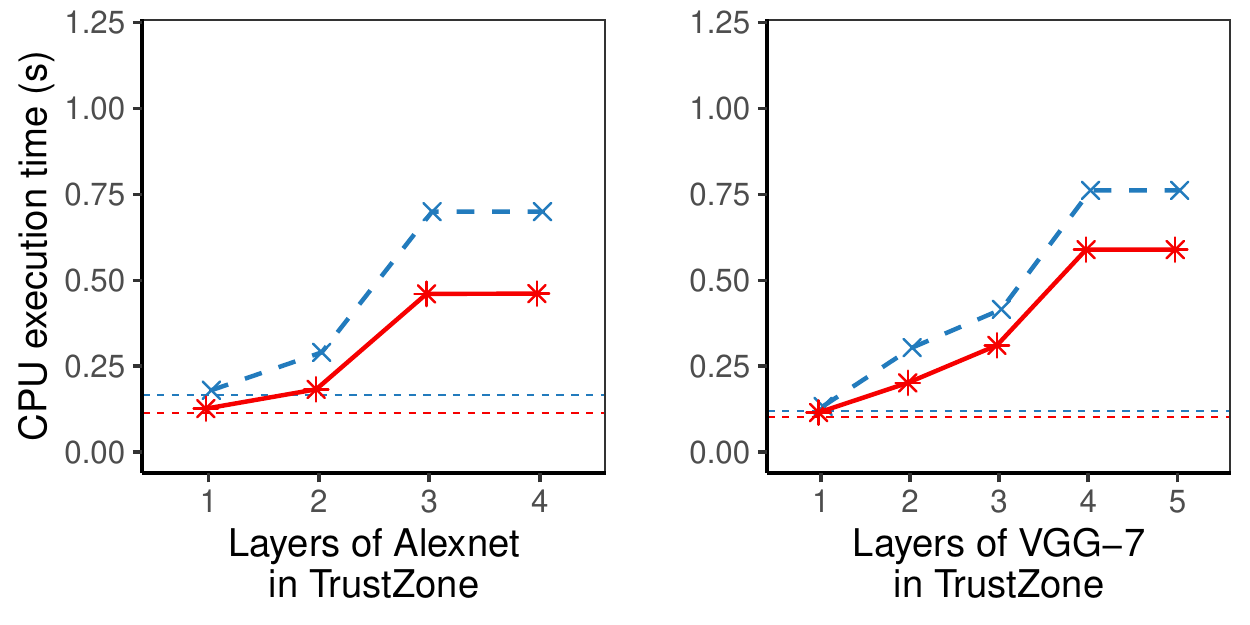}
     \caption{CPU time of inference}\label{fig:SM_exe_time_predict}
     \end{subfigure} %
    \caption{The CPU time of each step of training models or conducting inference on CIFAR-100 and ImageNet Tiny, protecting consecutive last layers using TrustZone (For example: when putting the last layers in the TrustZone, $1$ refers to the cost function and the softmax layer, $2$ includes $1$ and the previous fully-connected layer, $3$ includes $2$ and the previous convolutional layers, etc. Horizontal dashed lines
    (~\dottedred~and~\dottedblue~)
represent the baseline where all layers are out of the TrustZone. 20 times for each trial, and error bars are 95\% CI. Several error bars of data points are invisible as they are too small to be shown in this figure as well as the following figures).}
    \label{fig:CPU_cost}
\end{figure*}

%%%
\begin{figure}[t!]
    \centering
    \begin{subfigure}{0.5\columnwidth}
     \includegraphics[scale=0.66]{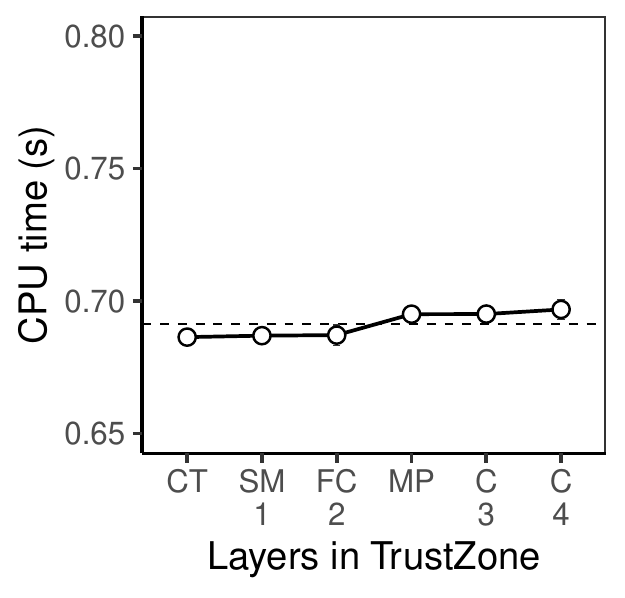}
     \caption{Training with Alexnet}\label{fig:cputime_train_alexnet_details}
   \end{subfigure}%
    \begin{subfigure}{0.5\columnwidth}
     \includegraphics[scale=0.66]{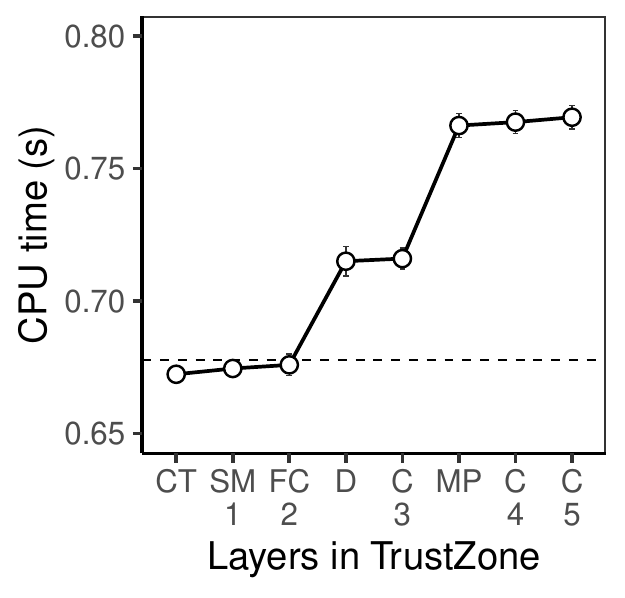}
     \caption{Training with VGG-7}\label{fig:cputime_train_vgg7_details}
     \end{subfigure}%
     \\
    \begin{subfigure}{0.5\columnwidth}
     \includegraphics[scale=0.66]{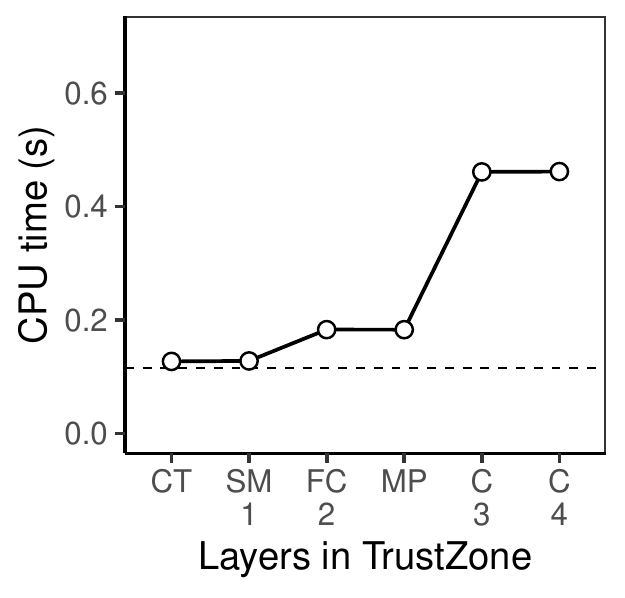}
     \caption{Inference with Alexnet}\label{fig:cputime_predict_alexnet_details}
   \end{subfigure}%
    \begin{subfigure}{0.5\columnwidth}
     \includegraphics[scale=0.66]{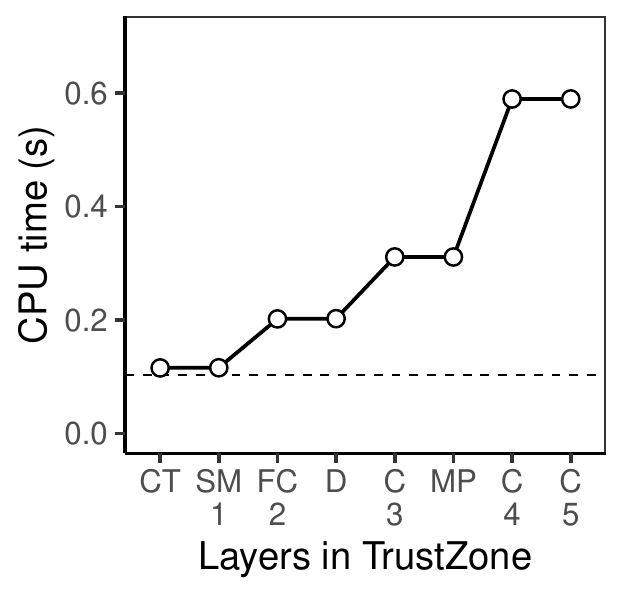}
     \caption{Inference with VGG-7}\label{fig:cputime_predict_vgg7_details}
     \end{subfigure}%
    \caption{The CPU time of each step of training models or conducting inference on CIFAR-100, protecting consecutive last layers using TrustZone (Note: The x-axis corresponds to several last layers included in the TrustZone. \emph{CT}, \emph{SM}, \emph{FC}, \emph{D}, \emph{MP}, and \emph{C} refer to the cost, softmax, fully connected, dropout, maxpooling, convolutional layers. 1, 2, 3, 4, and 5 in the x-axis are corresponding to the x-axis of Figure~\ref{fig:CPU_cost}. Horizontal dashed lines (~\dashedblack~) represent the baseline where all layers are out of the TrustZone. 20 times for each trial, and error bars are 95\% CI).}
    \label{fig:cputime_details}
\end{figure}

As shown in Figure~\ref{fig:CPU_cost}, the results indicate that including more layers in the TrustZone results in an increasing CPU time for deep learning operations, where the most expensive addition is to put the maximum number of layers. Figure~\ref{fig:SM_exe_time_train} shows the CPU time when \emph{training} AlexNet and VGG-7 with TrustZone on CIFAR-100 and ImageNet Tiny dataset, respectively. This increasing trend is significant and consistent for both datasets (CIFAR-100: $F_{(6,133)}=29.37, p<0.001$; $F_{(8,171)}=321.3, p<0.001$. ImageNet Tiny: $F_{(6,133)}=37.52, p<0.001$; $F_{(8,171)}=28.5, p<0.001$). We also observe that protecting only the last layer in the TrustZone has negligible effect on the CPU utilization, while including more layers to fully utilize the TrustZone during training can increase CPU time (by 10\%). For inference, the increasing trend is also significant (see Figure~\ref{fig:SM_exe_time_predict}). It only increases CPU time by around 3\% when protecting only the last layer which can increase up to $10\times$ when the maximum possible number of layers is included in the TrustZone.
%%%
\begin{figure}[t!]
    \centering
    \begin{subfigure}{1\columnwidth}
     \includegraphics[scale=0.66]{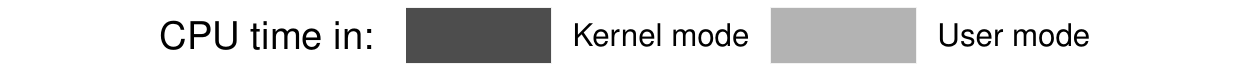}
   \end{subfigure}%
   \\
    \begin{subfigure}{0.5\columnwidth}
     \includegraphics[scale=0.66]{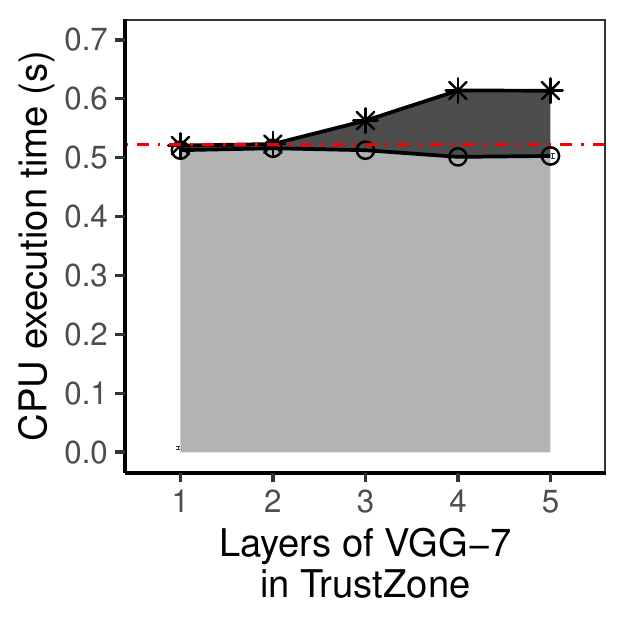}
     \caption{Training on CIFAR-100}\label{fig:SM_exe_time_train_details}
     \end{subfigure}%
    \begin{subfigure}{0.5\columnwidth}
     \includegraphics[scale=0.66]{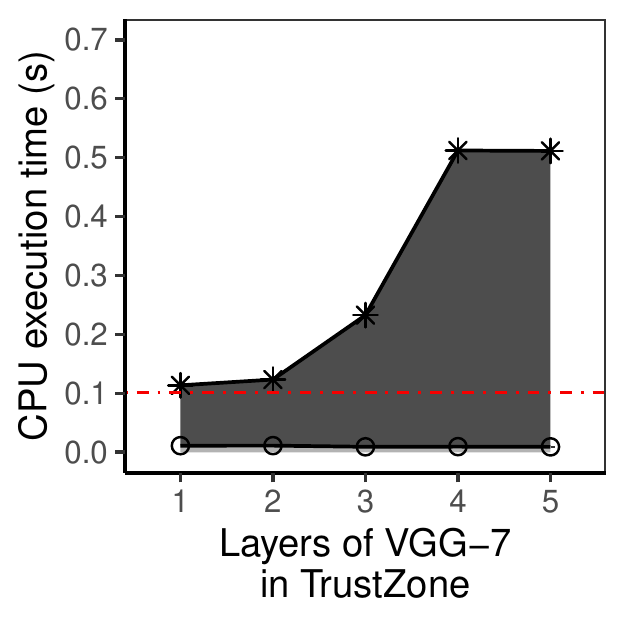}
     \caption{Inference on CIFAR-100}\label{fig:SM_exe_time_inference_details}
     \end{subfigure}%
    \caption{The CPU execution time in user mode and kernel mode of each step of training the model or conducting inference on CIFAR-100, protecting consecutive last layers using TrustZone (Note: Horizontal dot-dashed lines (~\dashedred~) represent the baseline where all layers are out of the TrustZone. 20 times for each trial. CPU time in user mode in Figure \ref{fig:SM_exe_time_inference_details} is too small to be shown).}
    \label{fig:cputime_kernel_user}
\end{figure}

To further investigate the increasing CPU execution time effect, we analyzed all types of layers (both \emph{trainable} and \emph{non-trainable}) separately in the TrustZone. Trainable layers have parameters (e.g., weights and biases) that are updated (i.e., trainable) during the training phase. Fully connected layers and convolutional layers are trainable. Dropout, softmax, and maxpooling layers are non-trainable. As shown in Figure~\ref{fig:cputime_details}, different turning points exist where the CPU time significantly increases ($p<0.001$) compared to the previous configuration (i.e.,~one more layer is moved into the TrustZone) (Tukey HSD~\cite{abdi2010tukey} was used for the post hoc pairwise comparison). When conducting \emph{training}, the turning points appear when putting the maxpooling layer in the TrustZone for AlexNet (see Figure~\ref{fig:cputime_train_alexnet_details}) and when putting the dropout layer and the maxpooling layer for VGG-7 (see Figure~\ref{fig:cputime_train_vgg7_details}). All these layers are non-trainable. When conducting \emph{inference}, the turning points appear when including the convolutional layers in TrustZone for both AlexNet (see Figure~\ref{fig:cputime_predict_alexnet_details}) and VGG-7 (see Figure~\ref{fig:cputime_predict_vgg7_details}), which are one step behind those points when conducting training.

One possible reason for the increased CPU time during \emph{inference} is that the TrustZone needs to conduct extra operations (e.g.,~related secure memory allocation) for the trainable layer, as shown in Figure~\ref{fig:cputime_predict_alexnet_details} and Figure~\ref{fig:cputime_predict_vgg7_details} where all increases happen when one trainable layer is included in the TrustZone. Since we only conduct one-time inference during experiments, the operations of invoking TEE libraries, creating the TA, and allocating secure memory for the first time significantly increased the execution time compared to the next operations. Every subsequent inference attempt (continuously without rebuilding the model) does not include additional CPU time overhead. Figure~\ref{fig:cputime_kernel_user} also shows that most of the increased CPU execution time (from $\sim$0.1s to $\sim$0.6s) is observed in the kernel mode---which includes the execution in TrustZone. The operation that needs to create the TA (to restart the TEE and load TEE libraries from scratch), such as one-time inference, should be taken care of by \emph{preloading} the TA before conducting inference in practical applications.

During \emph{training}, the main reason for the increased CPU time is that protecting non-trainable layers in the TrustZone results in an additional transmission of their previous trainable layers from the REE to the TrustZone. Non-trainable layers (i.e.,~dropout and max-pooling layers) are processed using a trainable layer as the base, and the non-trainable operation manipulates its previous layer (i.e.,~the trainable layer) directly. To hide the non-trainable layer and to prevent its next layer from being transferred to the REE during backward propagation (as mentioned in Section~\ref{sec:dnn_partitioned_exe}), we also move the previous convolutional layer to the TrustZone, which results in the turning points of the training are one layer in front of the turning points during inference. Therefore, in practical applications, we should protect the trainable layer and its previous non-trainable layer together, since only protecting the non-trainable layer still requires moving its trainable layer into TrustZone and does not reduce the cost.
%%%
\begin{figure*}[t!]
    \centering
     \begin{subfigure}{1.13\columnwidth}
     \includegraphics[scale=0.65]{figs_tee_legends.pdf}
     \end{subfigure} %
     \\
    \begin{subfigure}{1\columnwidth}
     \includegraphics[scale=0.65]{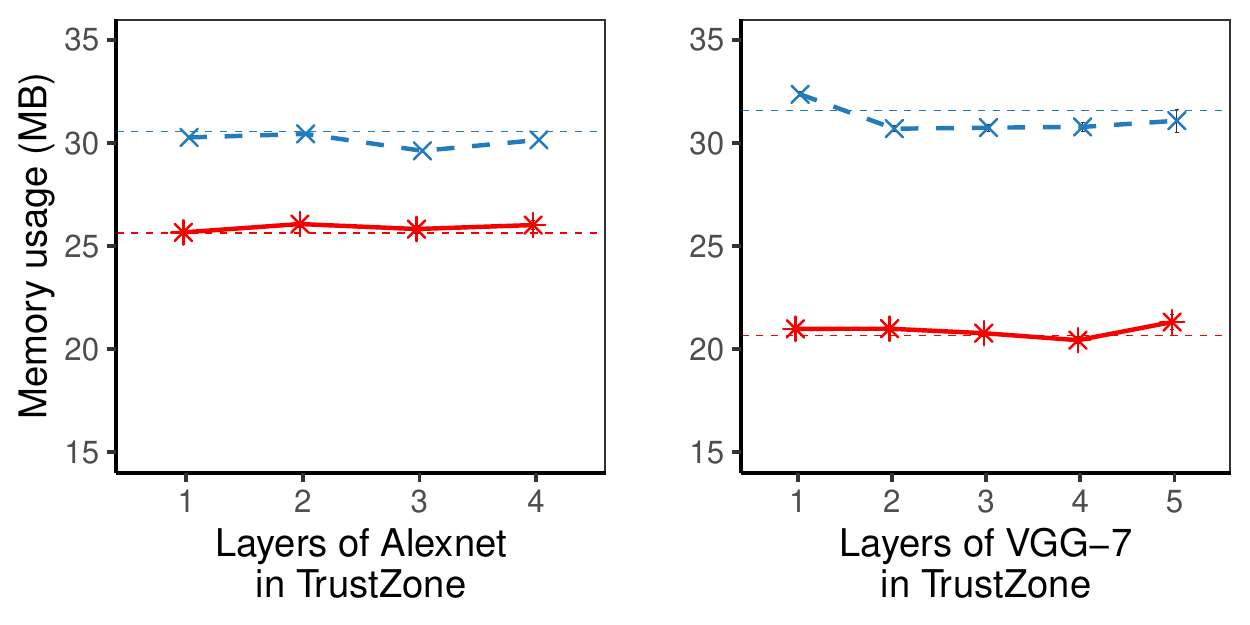}
     \caption{Memory usage of training}\label{fig:SM_mem_train}
   \end{subfigure}%
    \begin{subfigure}{1\columnwidth}
     \includegraphics[scale=0.65]{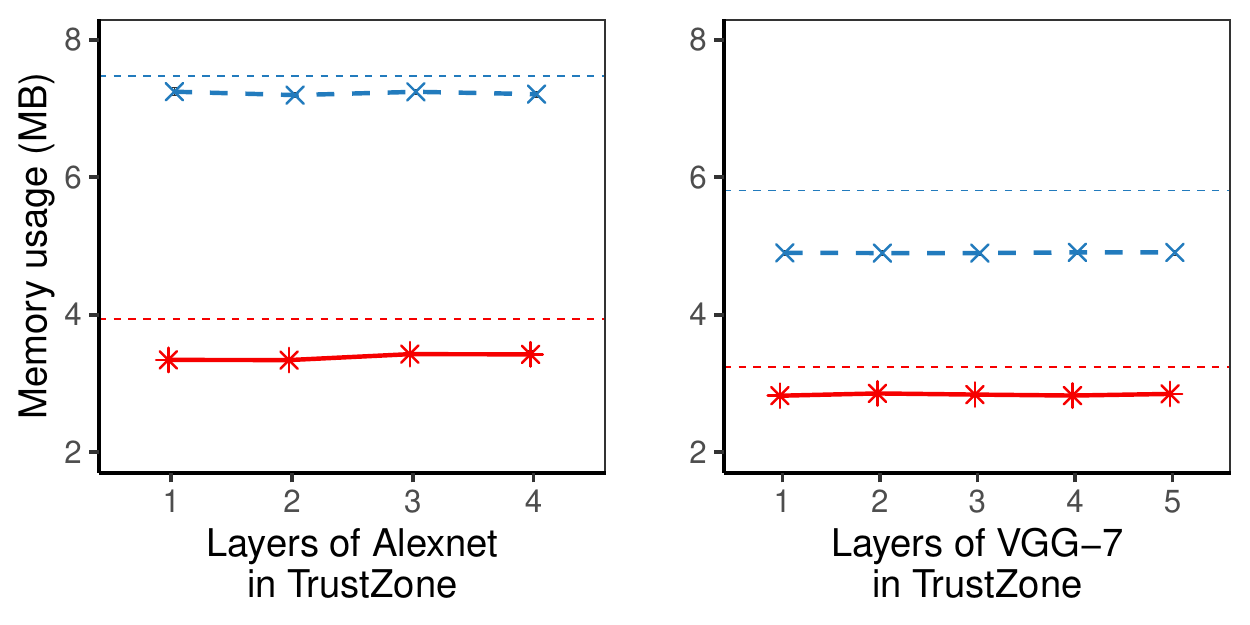}
     \caption{Memory usage of inference}\label{fig:SM_mem_predict}
     \end{subfigure}
    \\
    \begin{subfigure}{1\columnwidth}
     \includegraphics[scale=0.65]{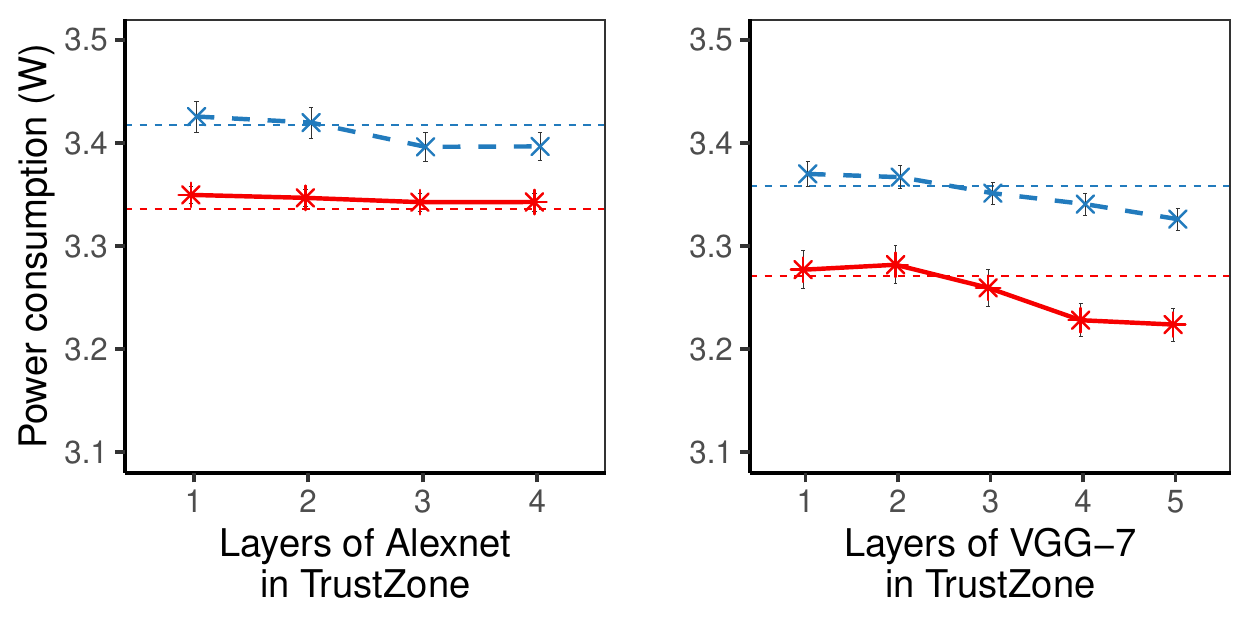}
     \caption{Power consumption of training}\label{fig:SM_power_train}
   \end{subfigure}%
    \begin{subfigure}{1\columnwidth}
     \includegraphics[scale=0.65]{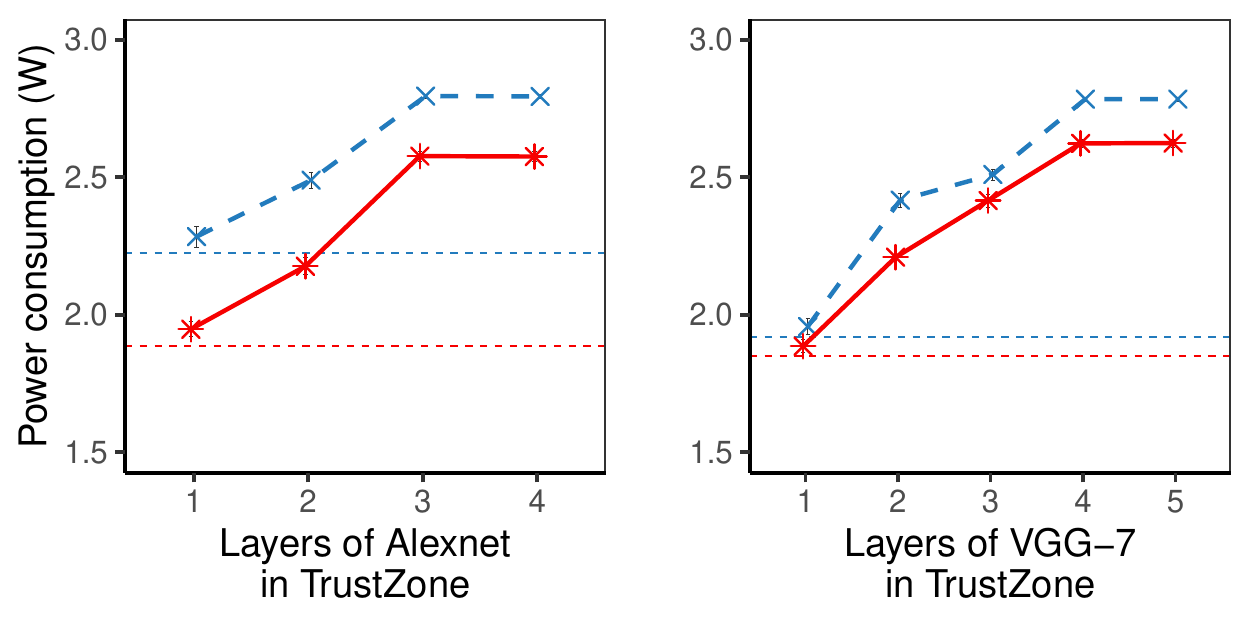}
     \caption{Power consumption of inference}\label{fig:SM_power_predict}
     \end{subfigure}
    \caption{The memory usage and power consumption of training models, while conducting training or inference on CIFAR-100 and ImageNet Tiny, protecting consecutive last layers using TrustZone (Note: Horizontal dashed lines (~\dottedred~and~\dottedblue~) %
represent the baseline where all layers are outside the TrustZone. 20 times for each trial, error bars are 95\% CI).}
    \label{fig:mem_power_cost}
\end{figure*}

%%%%%%%%%%%%%%  subsection  %%%%%%%%%%%%%%
\subsection{Memory Usage}

\emph{Training} with the TrustZone does not significantly influence the memory usage (in the REE) as it is similar to training without TrustZone (see Figure~\ref{fig:SM_mem_train}). \emph{Inference} with TrustZone uses less memory (in the REE) (see Figure~\ref{fig:SM_mem_predict}) but there is still no difference when more layers are placed into TrustZone. Memory usage (in the REE) \emph{decreases} since layers are moved to TrustZone and occupy secure memory instead. We measure the TA's memory usage using all mapping sizes in secure memory based on the TA's abort information. The TA uses five memory regions for sizes of \texttt{0x1000}, \texttt{0x101000}, \texttt{0x1e000}, \texttt{0xa03000}, and \texttt{0x1000} which is $11408 KiB$ in total for all configurations. The mapping size of secure memory is fixed when the TEE run-time allocates memory for the TA, and it does not influence when moving more layers into the memory. Therefore, because of the different model sizes, a good setting is to maximize the TA's memory mapping size in TrustZone in order to hold several layers of a possible large model.

%%%%%%%%%%%%%%  subsection  %%%%%%%%%%%%%%
\subsection{Power Consumption}

For \emph{training}, the power consumption significantly decreases (p < 0.001) when more layers are moved inside TrustZone (see Figure~\ref{fig:SM_power_train}). In contrast, the power consumption during \emph{inference} significantly increases (p < 0.001) as shown in Figure~\ref{fig:SM_power_predict}. In both training and inference settings, the trend of power consumption is likely related to the change of CPU time (see Figure~\ref{fig:CPU_cost}). More specifically, trajectories of them in figures have the same turning points (i.e., decreases or increases when moving the same layer to the TEE). One reason for the increased power consumption during inference is the significant increase in the number of CPU executions for invoking the required TEE libraries that consume additional power. When a large number of low-power operations (e.g.,~memory operations for mapping areas) are involved, the power consumption (i.e., energy consumed per unit time) could be lower compared to when a few CPU-bound computationally-intensive operations are running. This might be one of the reasons behind the decreased power consumption during training.
%%%
\begin{figure}[t!]
    \centering
    \includegraphics[width=\linewidth]{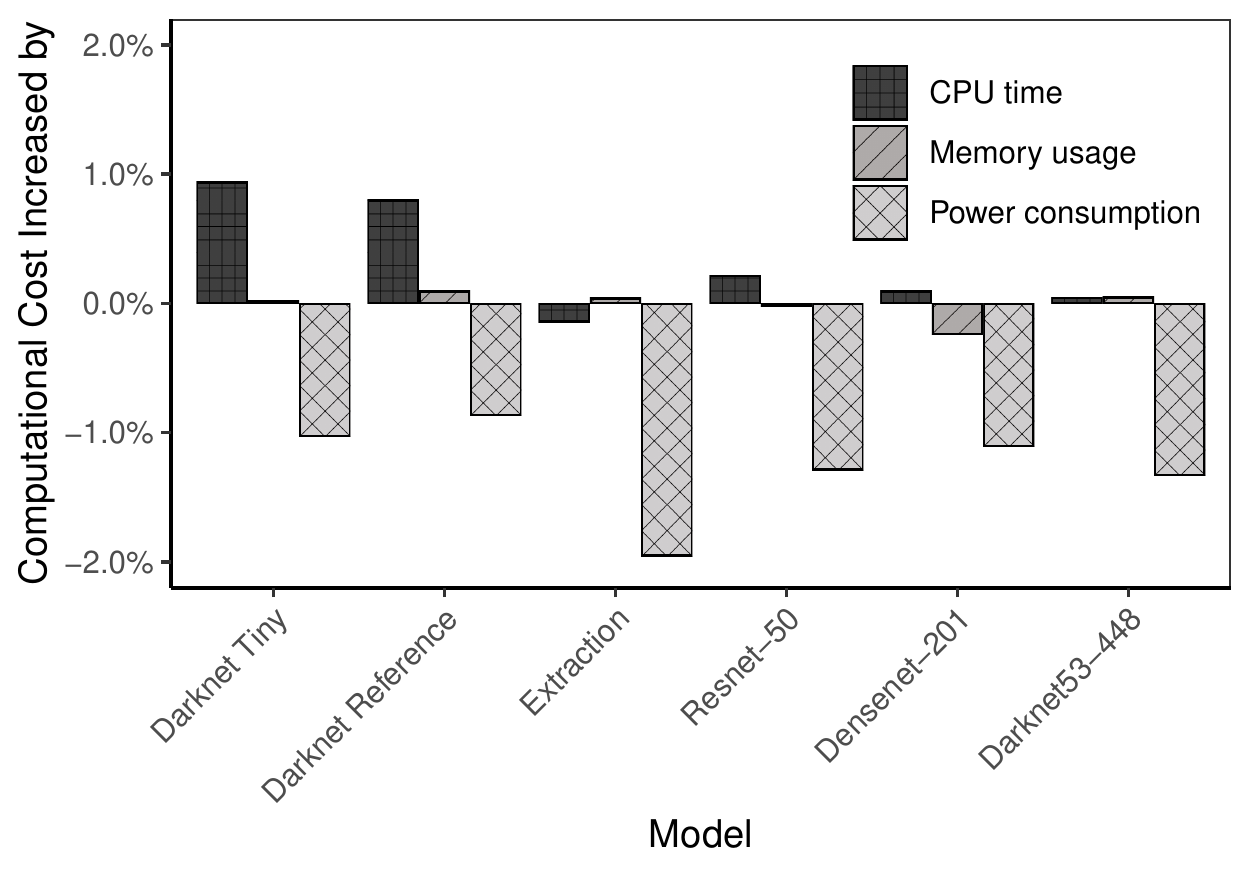}
    \caption{Performance on protecting the last layer of models trained on ImageNet in TrustZone for inference (Note: 20 times per trial; error bars are too small to be visible in the plot).}
    \label{fig:large_model}
\end{figure}
%%%

\vspace{5pt}\noindent \textbf{System performance on large models}.
We also test the performance of \legendsystem{} on several models trained on ImageNet when protecting the last layer only, including the softmax layer (or the pooling layer) and the cost layer in TrustZone, in order to hide confidence scores and the calculation of cost. The results show that the overhead of protecting large models is negligible (see Figure~\ref{fig:large_model}): increases in CPU time, memory usage, and power consumption are lower than $2\%$ for all models. Among these models, the smaller models (e.g.,~Tiny Darknet and Darknet Reference model) tend to have a higher rate of increase of CPU time compared to the larger models (e.g.,~Darknet-53 model), indicating that with larger models, the influence of TrustZone protection on resource consumption becomes relatively less.

\vspace{5pt}\noindent \textbf{System performance summary}.
In summary, \emph{it is practical to process a sequence of sensitive DNN model’s layers inside the TEE of a mobile device}. Putting the last layer in the TrustZone does not increase CPU time and only slightly increases memory usage (by no more than 1\%). The power consumption increase is also minor (no more than 0.5\%) when fine-tuning the models. For inference, securing the last layer does not increase memory usage but increases CPU time and power consumption (by 3\%). Including more layers to fully utilize the TrustZone during training can further increase CPU time (by 10\%) but does not harm power consumption. One-time inference with multiple layers in the TrustZone still requires further development, such as utilizing preliminary load of the TA, in practical applications.

%%%%%%%%%%%%%%  subsection  %%%%%%%%%%%%%%
\subsection{Privacy}
\label{sec:privacy_meas_results}

%%%
\begin{figure*}[t!]
    \centering
     \begin{subfigure}{1\columnwidth}
     \includegraphics[scale=0.6]{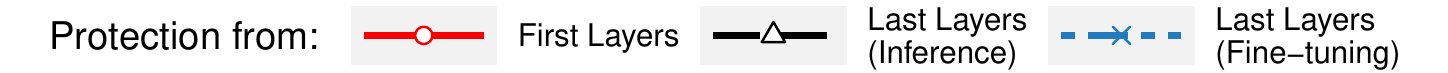}
     \end{subfigure} %
     \\
    \begin{subfigure}{1\columnwidth}
     \includegraphics[scale=0.65]{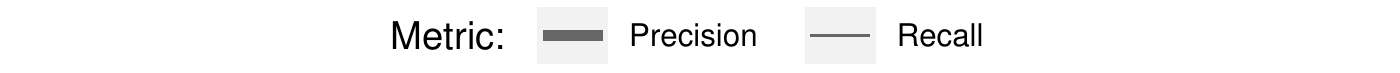}
     \end{subfigure}
     \\
    \begin{subfigure}{0.66\columnwidth}
     \includegraphics[height=5cm]{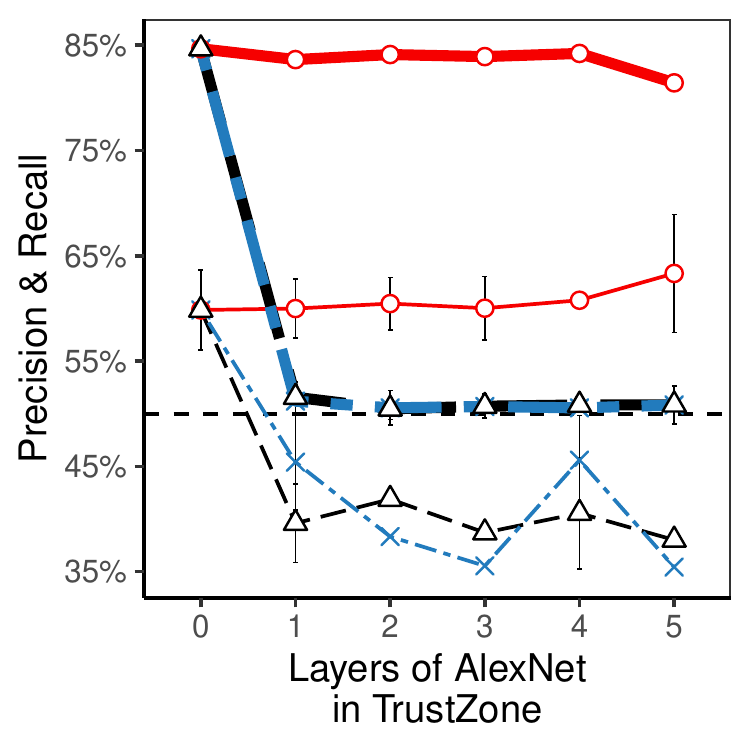}
     \label{fig:attack_alexnet_100cifar}
     \end{subfigure}
    \begin{subfigure}{0.66\columnwidth}
     \includegraphics[height=5cm]{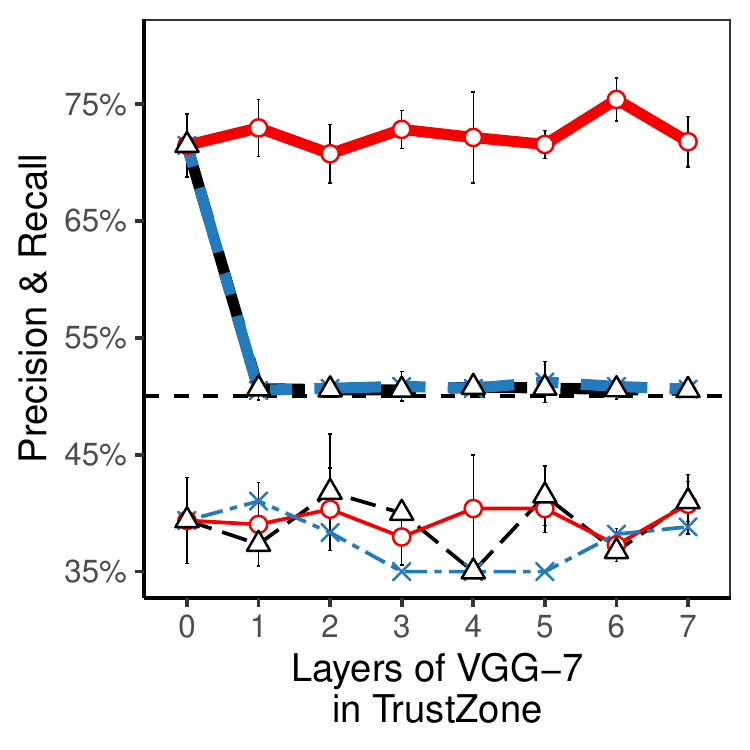}
     \label{fig:attack_vggnet_100cifar}
   \end{subfigure}
    \begin{subfigure}{0.66\columnwidth}
     \includegraphics[height=5cm]{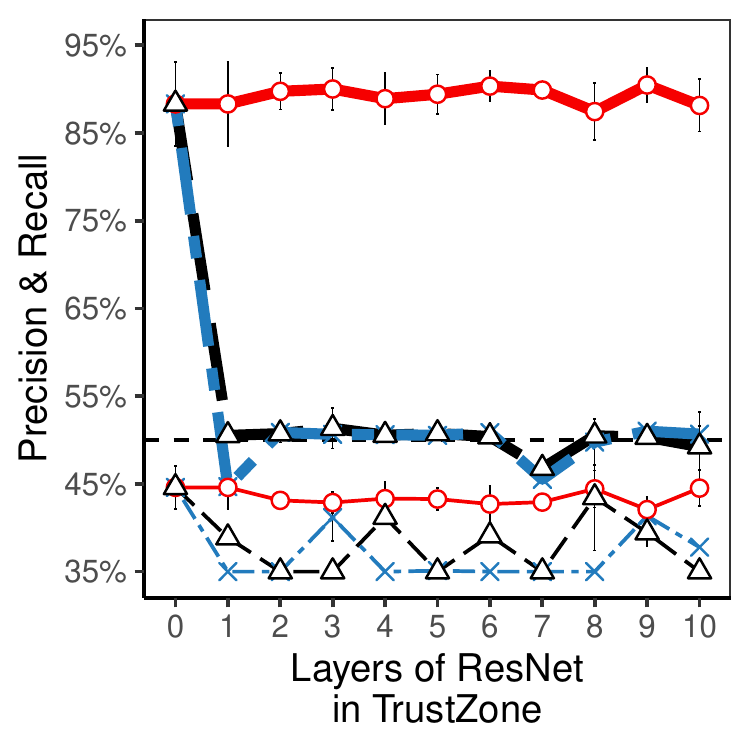}
     \label{fig:attack_resnet_100cifar}
   \end{subfigure}
    \caption{Precision and recall of white-box membership inference attacks when first or last layers of the model, trained on CIFAR-100, are protected using TrustZone. (Note: For first layer protection, 1 refers to the first layer, 2 refers to the first and the second layer, etc. For last layer protection, 1 refers to the last layer (i.e.,~the output layer), 2 refers to the last and second last layer, etc. 0 means that all layers are out of the TrustZone. Dashed lines at 50\% represent baselines (i.e.,~random guess). Each trial has been repeated 5 times, and error bars are 95\% CI).}
    \label{fig:privacy_attack_results}
\end{figure*}

\begin{table}[t!]
\small
\caption{Training and testing accuracy (Acc.) and corresponding MIA precision (Pre.) with or without \legendsystem{} (DTZ) of all models and datasets.}
\label{tab:accuracy}
\resizebox{\linewidth}{!}{%
\begin{tabular}{|l|l|l|l|l|l|}
\hline
Dataset & Model  & \makecell{Train \\ Acc.}  & \makecell{Test \\ Acc.} & \makecell{Attack \\ Pre.} & \makecell{Attack \\ Pre.\\(DTZ)} \\ \hline
\multirow{3}{*}{\makecell{CIFAR\\-100}} &AlexNet  &97.0\%    &43.9\%    &84.7\%    &51.1\% \\ \cline{2-6} 
                           &VGG-7    &83.8\%    &62.7\%    &71.5\%    &50.5\%     \\ \cline{2-6} 
                           &ResNet-100   &99.6\%    &72.4\%    &88.3\%    &50.6\%     \\ \hline
\multirow{3}{*}{\makecell{ImageNet\\Tiny}} &AlexNet  &40.3\%    &31.5\%   &56.7\%   &50.0\%  \\ \cline{2-6} 
                                           &VGG-7    &57.1\%    &48.6\%   &54.2\%   &50.8\%  \\ \cline{2-6} 
                                           &ResNet-110   &62.1\%    &54.2\%   &54.6\%   &50.2\%  \\ \hline
\end{tabular}
}
\end{table}

We conduct the white-box MIA (Section~\ref{sec:measure_mia}) on all target models (see Section~\ref{sec:model_dataset} for the choice of models) to analyze the privacy risk while protecting several layers in the TrustZone. We used the standard \emph{precision} and \emph{recall} metrics, similar to previous works~\cite{shokri2017membership}. In our context, precision is the fraction of records that an attacker infers as being members, that are indeed members in the training set. Recall is the fraction of training records that had been identified correctly as members. The performance for both models and MIAs are shown in Table~\ref{tab:accuracy}. Figure~\ref{fig:privacy_attack_results} shows the attack success precision and recall for different configurations of \legendsystem. In each configuration, a different number of layers is protected by TrustZone before we launch the attack. The configurations with zero layers protected correspond to \legendsystem{} being disabled (i.e.,~with our defense disabled). In particular, we measure the MIA adversary's success following two main configuration settings of \legendsystem. In the first setting, we incrementally add consecutive layers in the TrustZone starting from the front layers and moving to the last layers until the complete model is protected. In the second setting we do the opposite: we start from the last layer and keep adding previous layers in TrustZone for each configuration. Our results show that when protecting the first layers in TrustZone, the attack success precision does not change significantly. In contrast, hiding the last layers can significantly decrease the attack success precision, even when only a single layer (i.e.,~the last layer) is protected by TrustZone. The precision decreases to $\sim$50\% (random guessing) no matter how accurate the attack is before the defense. For example, for the AlexNet model trained on CIFAR-100, the precision drops from 85\% to $\sim$50\% when we only protect the \emph{last} layer in TrustZone. Precision is much higher than recall since the number of \emph{members} in the adversary's training set is larger than that of \emph{non-members}, so the MIA model predicts \emph{member} images better. The results also show that the membership information that leaks during inference and fine-tuning is very similar. Moreover, according to~\cite{nasr2018comprehensive} and~\cite{shokri2017membership}, the attack success precision is influenced by the size of the attackers' training dataset. We used relatively large datasets (half of the target datasets) for training MIA models so that it is hard for the attacker to increase success precision significantly in our defense setting. Therefore, by hiding the last layer in TrustZone, the adversary's attack precision degrades to 50\% (random guess) while the overhead is under 3\%.

%%%
\begin{figure}[t!]
    \centering
    \begin{subfigure}{1\columnwidth}
     \includegraphics[scale=0.65]{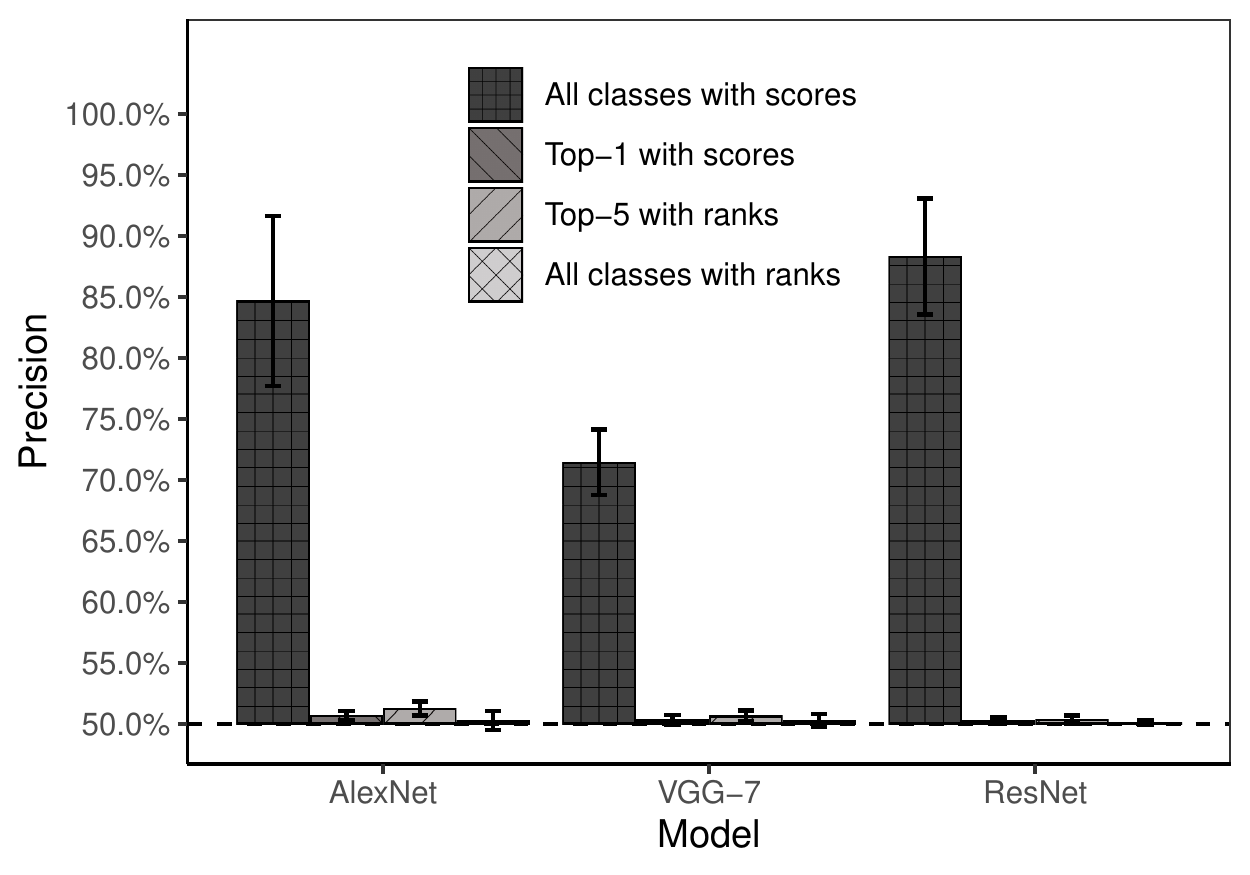}
     \label{fig:attack_control_outputs}
   \end{subfigure}%
    \caption{Precision of white-box membership inference attacks on models trained on CIFAR-100 when only outputs are protected using TrustZone (Dashed lines at 50\% represent baselines (i.e.,~random guess). 5 times for each trial, and error bars are 95\% CI).}
    \label{fig:attack_output_control}
\end{figure}

We also evaluated the privacy risk when \legendsystem{} protects the model's outputs in TrustZone by \emph{normalizing} it before outputting prediction results. In this configuration we conduct the white-box MIAs when all other layers (in the untrusted REE) are accessible by the adversary. This means that the cost function is protected, and the confidence score's outputs are controlled by TrustZone. Three combinations of models and datasets, including AlexNet, VGG-7, and ResNet on CIFAR-100 are selected as they were identified as more \emph{vulnerable} (i.e.,~with high attack precision see Table~\ref{tab:accuracy}) to MIAs~\cite{nasr2018comprehensive}. \legendsystem{} is set to control the model's outputs in three different ways: (a) top-1 class with its confidence score; (b) top-5 classes with their confidence scores; (c) all classes with their confidence scores. As shown in Figure~\ref{fig:attack_output_control} all three methods can significantly ($p<0.001$) decrease the attack success performance to around 50\% (i.e.,~random guess). Therefore, \emph{we found that it is highly practical to use \legendsystem{} to tackle MIAs: it incurs low resource consumption cost while achieving high privacy guarantees}.
\section{Discussion}
\label{sec:discussion}

%%%%%%%%%%%%%%  subsection  %%%%%%%%%%%%%%
\subsection{System Performance}

\noindent\textbf{Effects of the model size.} We showed that protecting large models with TrustZone tends to have a lower rate of increase of CPU execution time than protecting small models (see Figure~\ref{fig:large_model}). One possible explanation is that the last layer of a larger model uses a lower proportion of computational resources in the whole model compared to that of a smaller model. We have also examined the effect of different hardware: we executed our implementation of DarkneTZ with similar model sizes on a Raspberry Pi 3 Model B (RPi3B) and found it to have a lower rate of increase of cost (i.e.,~lower overhead) than when executed on the Hikey board~\cite{mo2019towards}. This is because the Hikey board has much faster processors optimized for matrix calculations, which renders additional operations of utilizing TrustZone more noticeable compared to other normal executions (e.g.,~deep learning operations) in the REE. Moreover, our results show that a typical configuration (16MiB secure memory) of the TrustZone is sufficient to hold at least the last layer of practical DNN models (e.g., trained on ImageNet). However, it is challenging to fit multiple layers of large models in a significantly smaller TEE. We tested a TEE with 5MiB secure memory on a Grapeboard\footnote{\url{https://www.grapeboard.com/}}: only 1,000 neurons (corresponding to 1,000 classes) in the output layer already occupy 4MiB memory when using floating-point arithmetic. In such environments, model compression, such as pruning~\cite{han2015deep} and quantization~\cite{wang2019haq, jacob2018quantization}, could be one way to facilitate including more layers in the TEE. Lastly, we found that utilizing TEEs for protecting the last layer does not necessarily lead to resource consumption overhead, which deserves further investigation in future work. Overall, our results show that utilizing TrustZone to protect outputs of large DNN models is effective and highly efficient.

\vspace{5pt}\noindent\textbf{Extrapolating for other mobile-friendly models.} We have used Tiny Darknet and Darknet Reference for testing \legendsystem's performance on mobile-friendly models (for ImageNet classification). Another widely-used DNNs on mobile devices, Squeezenet~\cite{iandola2016squeezenet} and Mobilenet~\cite{howard2017mobilenets}, define new types of convolutional layers are not supported in Darknet framework currently. We expect these to have a similar privacy and TEE performance footprint because of the comparable size of model (4MB, 28MB, 4.8MB, 3.4MB for Tiny Darknet, Darknet Reference, Squeezenet, and Mobilenet, respectively), floating-point operations (980M, 810M, 837M, 579M), and model accuracy (58.7\%, 61.1\%, 59.1\%, and 71.6\% for Top-1)\footnote{\url{https://github.com/albanie/convnet-burden} and \url{https://pjreddie.com/darknet/tiny-darknet/}}. 

\vspace{5pt}\noindent\textbf{Improving performance.}  Modern mobile devices usually are equipped with GPU or specialized processors for deep learning such as NPU. Our current implementation only uses the CPU but can be extended to utilizing faster chips (i.e.,~GPU) by moving the first layers of the DNN that is always in the REE to these chips. By processing several layers of a DNN in a TEE (SGX) and transfer all linear layers to a GPU, Tramer et al.~\cite{tramer2018slalom} have obtained 4x to 11x increase for verifiable and private inference in terms of VGG16, MobileNet, and ResNet. For edge devices, another way for expediting the deep learning process is to utilize TrustZone's AXI bus or peripheral bus, which also has an additional secure bit on the address. Accessing a GPU (or NPU) through the secure bus enables the TrustZone to control the GPU so that the confidentiality of DNN models on the GPU cannot be breached and achieve faster executions for partitioned deep learning on devices.

%%%%%%%%%%%%%%  subsection  %%%%%%%%%%%%%%
\subsection{Models' Privacy}

\noindent\textbf{Defending against other adversaries.} \legendsystem{} is not only capable of defending MIAs by controlling information from outputs, but also capable of defending other types of attacks such as training-based model inversion attack~\cite{fredrikson2015model, yang2019neural} or GAN attack~\cite{hitaj2017deep} as they also highly depend on the model's outputs. In addition to that, by controlling the output information during inference, \legendsystem{} can provide different privacy settings depending on different privacy policies to servers correspondingly. For example, options included in our experiments are outputting Top-1 only with its confidence scores, outputting Top-5 with their ranks, or outputting all classes with their ranks which all achieve strong defense against MIAs. Recent research~\cite{jia2019memguard} also manipulates confidence scores (i.e.,~by adding noises) to defend against MIAs, but their protection can be broken easily if the noise addition process is visible to the adversaries for a compromised OS. \legendsystem{} also protects layers while training models and conducting inference. The issue of private information leaked from layers' gradients becomes more serious considering that DNN models' gradients are shared and exchanged among devices in collaborated/federated learning. \cite{melis2019exploiting}'s work successfully shows private (e.g.,~membership) information about participants' training data using their updated gradients. Recent research~\cite{chou2019deep} further reveals that it is possible to recover images and texts from gradients in pixel-level and token-level, respectively, and the last layers have a low loss for the recovery. By using \legendsystem{} to limit information exposure of layers, this type of attack could be weakened.

\vspace{5pt}\noindent\textbf{Preserving model utility.} By "hiding" (instead of obfuscating) parts of a DNN model with TrustZone, \legendsystem{} preserves a model's privacy without reducing the utility of the model. Partitioning the DNN and moving its more sensitive part into an isolated TEE maintains its prediction accuracy, as no obfuscating technique (e.g.,~noise addition) is applied to the model. As one example of obfuscation, applying differential privacy can decrease the prediction accuracy of the model~\cite{yu2019differentially}. Adding noises to a model with three layers trained on MNIST leads to the model accuracy drop by $5\%$ for small noise levels ($\epsilon=8$) and by $10\%$ for large noise levels ($\epsilon=2$)~\cite{tensorflow2019privacy, abadi2016deep}. The drop increases to around $20\%$ for large level noises when training on CIFAR-10~\cite{abadi2016deep}. To obtain considerable accuracy when using differential privacy, one needs to train the model with more epochs, which is challenging for larger models since more computational resources are needed. In recent work, carefully crafted noise is added to confidence scores by applying adversarial examples~\cite{jia2019memguard}. Compared to the inevitable decreasing utility of adding noise, \legendsystem{} achieves a better trade-off between privacy and utility compared to differential privacy.
\section{Conclusion}
\label{sec:conclusion}

We demonstrated a technique to improve model privacy for a deployed, pre-trained DNN model using on-device Trusted Execution Environment (TrustZone). We applied the protection to individual sensitive layers of the model (i.e.,~the last layers), which encode a large amount of private information on training data with respect to Membership Inference Attacks. We analyzed the performance of our protection on two small models trained on the CIFAR-100 and ImageNet Tiny datasets, and six large models trained on the ImageNet dataset, during training and inference. Our evaluation indicates that, despite memory limitations, the proposed framework, \emph{\legendsystem}, is effective in improving models' privacy at a relatively low performance cost. Using \emph{\legendsystem{}} adds a minor overhead of under $3\%$ for CPU time, memory usage, and power consumption for protecting the last layer, and of $10\%$ for fully utilizing a TEE's available secure memory to protect the maximum number of layers (depending on the model size and configuration) that the TEE can hold. We believe that \emph{\legendsystem{}} is a step towards stronger privacy protection and high model utility, without significant overhead in local computing resources.

\section*{Acknowledgments}
We acknowledge the constructive feedback from the anonymous reviewers. Katevas and Haddadi were partially supported by the EPSRC Databox and DADA grants (EP/N028260/1, EP/R03351X/1). This research was also funded by a gift from Huawei Technologies, a generous scholarship from the Chinese Scholarship Council, and a hardware gift from Arm.

\balance
\bibliographystyle{ACM-Reference-Format}
\bibliography{00_main}

\end{document}